\documentclass[a4paper,conference]{IEEEtran}

\ifCLASSINFOpdf

\else

\fi

\usepackage[utf8]{inputenc} % allow utf-8 input
\usepackage[T1]{fontenc}    % use 8-bit T1 fonts
\usepackage{hyperref}       % hyperlinks
\usepackage{url}            % simple URL typesetting
\usepackage{booktabs}       % professional-quality tables
\usepackage{amsfonts}       % blackboard math symbols
\usepackage{nicefrac}       % compact symbols for 1/2, etc.
\usepackage{microtype}      % microtypography
\usepackage{lipsum}
\usepackage{graphicx}
\usepackage{multirow}
\usepackage{diagbox}
\usepackage{xcolor}

\usepackage{amssymb}% http://ctan.org/pkg/amssymb
\usepackage{pifont}% http://ctan.org/pkg/pifont
\newcommand{\cmark}{\ding{51}}%
\newcommand{\xmark}{\ding{55}}%

\begin{document}

% \title{Sequential 3D Hand Pose Estimation with Transformer}

\title{Learning Sequential Contexts using Transformer for 3D Hand Pose Estimation}

% \title{Learning Temporal and Angular Contexts using Transformer for 3D Hand Pose Estimation}

% Hand Pose Estimation using Transformer for Time and View Sequence 

% Hand Pose Estimation using Transformer for Time and View Sequence 

\author{\IEEEauthorblockN{Leyla Khaleghi,~Joshua Marshall,~Ali Etemad}
\IEEEauthorblockA{Dept. ECE and Ingenuity Labs Research Institute\\
Queen's University\\
Kingston, ON, Canada\\
\{leyla.khaleghi, joshua.marshall, ali.etemad\}@queensu.ca}
}

% \author{\IEEEauthorblockN{}
% \IEEEauthorblockA{\\
% \\
% \\
% \\
% }
% }

% \and
% \IEEEauthorblockN{Joshua Marshall}
% \IEEEauthorblockA{Department of Electrical and\\Computer Engineering\\
%  Ingenuity Labs Research Institute\\
%  Queen's University\\
%  Kingston, ON, Canada\\
% Email: joshua.marshall@queensu.ca}
% \and
% \IEEEauthorblockN{Ali Etemad}
% \IEEEauthorblockA{Department of Electrical and\\Computer Engineering\\
%  Ingenuity Labs Research Institute\\
%  Queen's University\\
%  Kingston, ON, Canada\\
% Email: ali.etemad@queensu.ca}
% }

\maketitle

\begin{abstract}
3D hand pose estimation (HPE) is the process of locating the joints of the hand in 3D from any visual input. HPE has recently received an increased amount of attention due to its key role in a variety of human-computer interaction applications. Recent HPE methods have demonstrated the advantages of employing videos or multi-view images, allowing for more robust HPE systems. Accordingly, in this study, we propose a new method to perform Sequential learning with Transformer for Hand Pose (SeTHPose) estimation. Our SeTHPose pipeline begins by extracting visual embeddings from individual hand images. We then use a transformer encoder to learn the sequential context along time or viewing angles and generate accurate 2D hand joint locations. Then, a graph convolutional neural network with a U-Net configuration is used to convert the 2D hand joint locations to 3D poses. Our experiments show that SeTHPose performs well on both hand sequence varieties, temporal and angular. Also, SeTHPose outperforms other methods in the field to achieve new state-of-the-art results on two public available sequential datasets, STB and MuViHand. 

\end{abstract}


\IEEEpeerreviewmaketitle

\section{Introduction}
Various human-computer interaction applications including gaming, autonomous driving, automatic sign-language recognition, and others, rely upon the estimation of hand joint locations---i.e., hand pose estimation (HPE)---from input visual data \cite{9773085,doosti2019hand,chen2020survey}. Despite the considerable improvements in HPE methods with the help of deep learning \cite{supanvcivc2018depth,li2019survey}, many methods still struggle to deal with challenges such as
% a number of challenges still remain in this field. These challenges include 
occlusions (which could happen by a body part or by a part of the hand itself, called self-occlusion), large number of possible hand poses (large pose space), large number of hand shapes, various skin colors, and sharp camera viewpoints.

So far, the majority of HPE solutions focus on using single RGB images as inputs \cite{zimmermann2017learning,iqbal2018hand,doosti2020hope,hasson2019learning,zhang2019end,boukhayma20193d,hampali2021handsformer,huang2020hot,lin2021end,yang2019aligning,spurr2018cross,mueller2018ganerated,panteleris2018using,chen2020dggan,yang2019disentangling,li2020exploiting,Ge_2019_CVPR}, overlooking the information in the sequential structure of the hand in time and/or across different views. Thus far, only a few studies \cite{yang2020seqhand, cai2019exploiting,khaleghi2021multi} have considered the temporal relations of hand frames in videos to improve the performance of HPE. Similarly, only a few methods \cite{khaleghi2021multi,chen2021mvhm} have fused multi-view hand images to capture the geometric relations among different viewpoints and improve HPE performance. 
%Yet, despite such sequential information (time/view) often being ignored for HPE, we believe their effective learning and aggregation could lead to more robust 
% , compensate the lack of data that a view could have and result in more accurate estimation in comparison to estimating from a single image. 
% Hence considering any sequential relationship in neural network, significantly boosts HPE performance. 

\begin{figure}[!t]
\centerline{\includegraphics[width=0.48\textwidth]{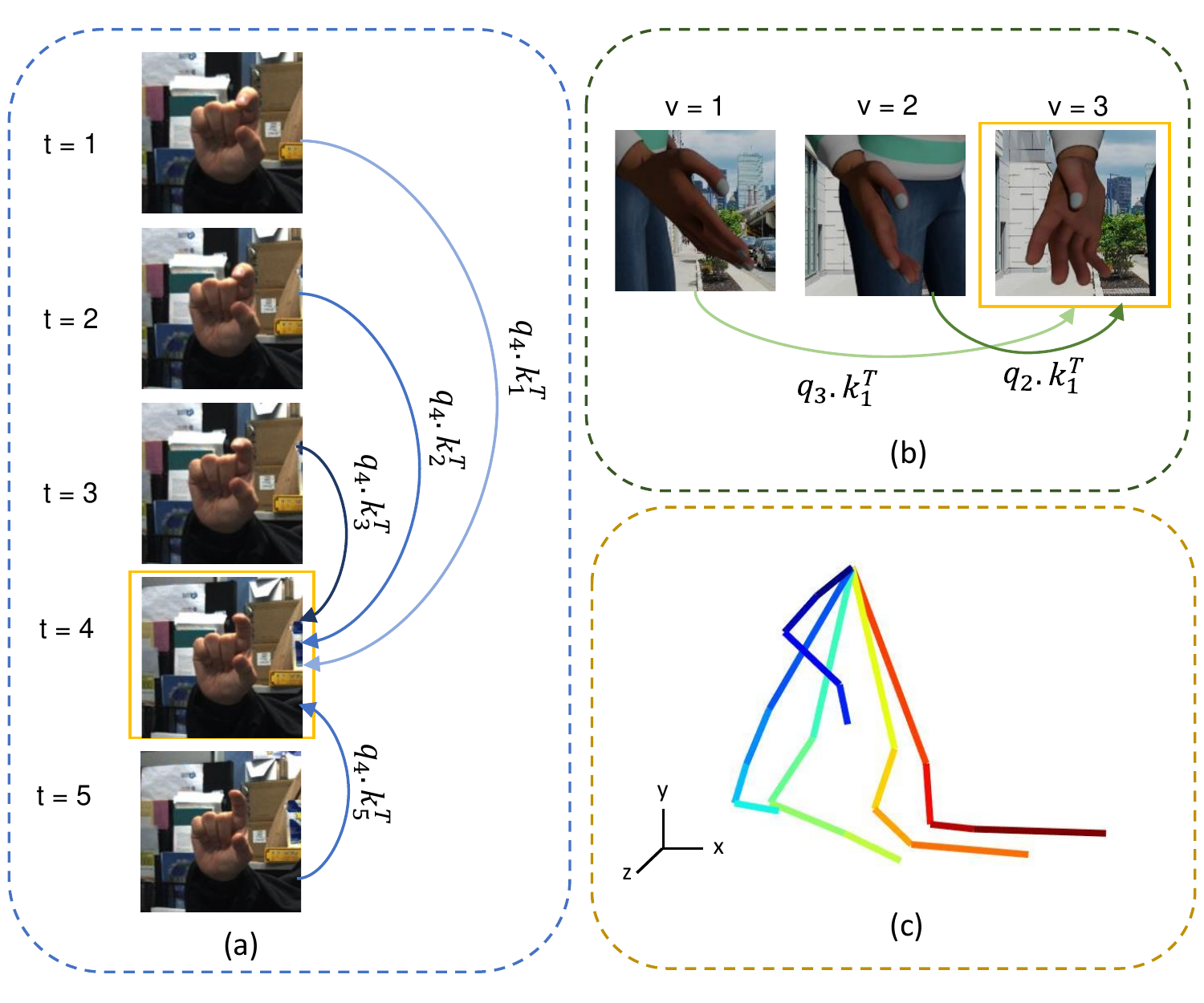}}
\caption {SeTHPose scores the relevancy of each pair of hand images in a sequence (time/view) according to its surrounding context with a self-attention mechanism to estimate the corresponding 3D hand pose.}
\label{fig:intro}
\end{figure}

Despite sequential information (time/view) often being ignored for HPE, we believe their effective learning and aggregation could lead to more robust contextualized representations and ultimately more accurate HPE. While recurrent neural networks have been used in the past for learning such time/view contextual information \cite{khaleghi2021multi,yang2020seqhand}, the Transformer self-attention \cite{vaswani2017attention}, which has recently
% which was initially developed in natural language processing for encoding the correlation between word embedding \cite{vaswani2017attention}, 
shown remarkable results in various computer vision tasks \cite{dosovitskiy2020image,khan2021transformers,han2020survey}, 
% The attention mechanism in this architecture 
appears to be a strong candidate for learning sequential information across time or views for enhancing 3D HPE. 
As a result, we propose a new HPE method called \underline{Se}quential learning with \underline{T}ransformer for \underline{H}and \underline{P}ose (SeTHPose) estimation. 
% **** Our model uses a combination of convolutional neural networks and a transformer encoder to learn the temporal or angular relationships in hand sequences. 
%A hand sequence could be consecutive frames of a video (temporal relationship) or multiple frames captured from multiple views at the same time (angular relationship).
%  Transformers self-attention \cite{vaswani2017attention} 
%  have emerged as a powerful solution in natural language processing, the attention mechanism in this architecture makes it a viable candidate for learning correlations across any sequential data \cite{khan2021transformers,han2020survey}, expanded to 3D HPE. As a result, we propose a new HPE method called \underline{Se}quential 
% \underline{T}ransformer \underline{Hand} pose estimation  \underline{Net}work (SeTHPose), based on a combination of convolutional neural networks and transformer self-attention which learn the angular or temporal relationships in hand sequence. It worth mentioning that in this research sequence is consider as consecutive frames of videos or multiple hand mages captured from multiple views simultaneously.
Our model first uses a CNN encoder to generate the spatial embeddings for each hand frame, followed by a transformer encoder to learn the contextual relations of the hand embeddings across time or views (Figures \ref{fig:intro}(a) and (b) demonstrate the different contextual data types, while (c) demonstrates a sample 3D pose output). Next, an MLP is used for estimating the 2D hand joint locations. This is followed by a graph convolutional U-network to convert the learned  2D structures to 3D hand joint locations. 
% We have employed the 2D HPE lifting to 3D HPE approaches through estimating depth information, instead of directly estimating the 3D hand pose from the input hand sequence. The short dependencies of the hand joint in the graph-based structure of the hand skeleton have been learned via a graph convolutional network (GCN), and finally the 3D hand poses are estimated. 
We validate our model on two publicly available \textit{sequential} datasets, STB \cite{zhang20163d} and MuViHand \cite{khaleghi2021multi}, and achieve state-of-the-art results by outperforming previous works in the area. Moreover, we perform additional experiments, such as ablation studies, to evaluate the impact of contextual learning within our model. While a few new techniques have been recently proposed to use transformers for HPE \cite{hampali2021handsformer,huang2020hot,lin2021end}, their use has been solely limited for obtaining better single-image embeddings or joint locations, while our method uses the transformer component to learn the contextual information across time or view sequences.
To the best of our knowledge, SeTHPose is the first method that learns \textit{sequential contextual information} with transformer for 3D HPE. %Our investigation demonstrates that SeTHPose is a practical approach in learning both angular/temporal relationships in 3D HPE. 

Our contributions in this paper are summarized as follows:
\begin{itemize}
\item  We propose SeTHPose, a method for strong 3D hand pose estimation. Our method relies on a transformer to learn contextual sequential information and produce better 2D joint locations, which are then converted to 3D using a graph convolutional U-Net. 
\item Our experiments show that SeTHPose works for two different types of sequential information, namely time and camera angles.
\item SeTHPose outperforms existing techniques in the area to achieve state-of-the-art results on two public datasets, STB and MuViHand.
\end{itemize}

We have organized the remainder of this paper as follows. A literature review on recent HPE methods is explained in Section 2. We introduce our proposed method in Section 3 in detail. Our experiments and results on SeTHPose are presented in Section 4. Finally, Section 5 provides the concluding remarks.

\section{Related Work}
In this study, we focus primarily on 3D HPE by exploiting sequential contexts; i.e., both temporal and angular. Accordingly in this section we review typical HPE techniques that rely on single images (non-sequential), as well as approaches that exploit either temporal and/or angular contexts (sequential). %although we will also review non-sequential HPE methods given the lack of comprehensive studies on sequential methods. Finally, we describe a few multi-view and temporal methods.

\subsection{Non-Sequential HPE}
In some prior HPE literature, estimating 2D hand locations had been used as an intermediary step for estimating 3D poses from RGB images. In \cite {zimmermann2017learning}, a 2D heatmap was generated directly from an image embedding for each hand joint, and a normalized 3D hand coordinate was then calculated from the 2D heatmap. Similarly, in \cite{iqbal2018hand}, the depth data was measured in addition to the initial 2D heatmaps, from which the 3D hand pose was calculated. In comparison to \cite {zimmermann2017learning} and \cite{iqbal2018hand}, \cite{doosti2020hope} is a two-stage approach to 3D hand pose estimation. The 2D joint locations were estimated rather than the 2D heatmaps, and the 3D hand pose was determined from the joint locations via a graph-based network.

The MANO hand model \cite{romero2017embodied} was employed in multiple HPE techniques for estimating hand poses. MANO maps two parameters, namely pose (joint angles) and shape (individual deformations of the hands), to a 3D hand mesh. In \cite{hasson2019learning}, MANO hand parameters were directly derived from the image embeddings obtained from a ResNet18. In \cite{zhang2019end}, MANO parameters were estimated based on 2D heatmaps and using hand mask supervision, calculated through hand vertices. In \cite{boukhayma20193d}, the MANO parameters were calculated by providing an RGB image and its corresponding 2D hand joints to a ResNet50.

A few HPE pipelines have employed the self-attention transformer mechanism. In \cite{hampali2021handsformer}, image embeddings along with 2D hand keypoint heatmaps were fed to a transformer in order to calculate the 3D hand positions. In \cite {huang2020hot}, hand images were fed to a ResNet encoder to generate hand joints in 2D, which were then passed through a transformer to estimate hand poses in 3D. 
% Lastly, the 3D hand pose is merged with the image embedding to generate MANO hand model parameters in order to estimate the hand mesh. 
In \cite{lin2021end}, encoded hand images were passed through a multilayer transformer in order to regress the 3D hand poses.

\subsection{Sequential HPE}
Exploiting sequential contextual information, either in the form of consecutive video frames or multi-view images, has rarely been studied in HPE literature.
In \cite{yang2020seqhand}, an LSTM layer followed image encoders to learn the temporal relationships between the consecutive hand images and generate the parameters for the MANO hand model for each frame. In \cite{cai2019exploiting} the temporal relationships between 2D hand joint locations were considered through GC layers with extra edges between the exact hand joints in the following frames, where state-of-the-art methods such as OpenPose \cite{simon2017hand} could be used to estimate the 2D hand joint locations. In \cite{chen2021mvhm}, a single-view HPE generated 3D camera coordinates for different views using multiple hand images. A graph-based structure then calculated the world coordinate by concatenating the camera coordinates. 

The only approach, in the literature that could learn both angular and temporal relationships for HPE is \cite{khaleghi2021multi}. In this paper, the 2D locations of the hand joints were generated by two LSTM layers that jointly learned the relevance of hand videos captured from multiple perspectives. By using GC layers that allowed for learning of hand joints' relevancy, the 3D hand pose could be estimated from the 2D hand joint locations. 

Despite the promising results obtained by transformers in other domains, there are currently no HPE approaches that examine the utility of transformer encoders for learning sequential contexts of hand data. This motivates our study where we present SeTHPose, the first sequential transformer-based 3D HPE method.

\section{Method}

\subsection{Problem Setup and Solution Overview}

% Although the initial Transformer self-attention was developed for encoding the correlation between word embedding, it has shown remarkable results in various computer vision tasks\cite{dosovitskiy2020image}. 
Our work is based on the idea of considering the temporal or angular contextual information in hand sequences as a practical solution for improving the performance of HPE \cite{yang2020seqhand, cai2019exploiting,khaleghi2021multi,chen2021mvhm}.
Accordingly, we can denote our proposed method SeTHPose using $\zeta$, which can predict 3D hand poses from the corresponding hand sequence $\varphi$, such that 
\begin{equation}
    \textit{P}= \zeta[\varphi], 
\end{equation}
where $\varphi$ is a sequence of RGB hand frames $\in \mathbb{R}^{{N} \times {3} \times{H} \times{W}}$, $N$ is the sequence length, $H$ is the image height, and $W$ is the image width. Moreover, $P\in \mathbb{R}^{{N} \times {j} \times {3}}$ is the sequence of corresponding 3D hand poses that could describe $N$ hand skeletons, each consisting of $j$ joints. A sample output 3D hand pose can be seen in Figure \ref{fig:intro}(c).

\begin{figure*}[!t]
\centerline{\includegraphics[width=0.95\textwidth]{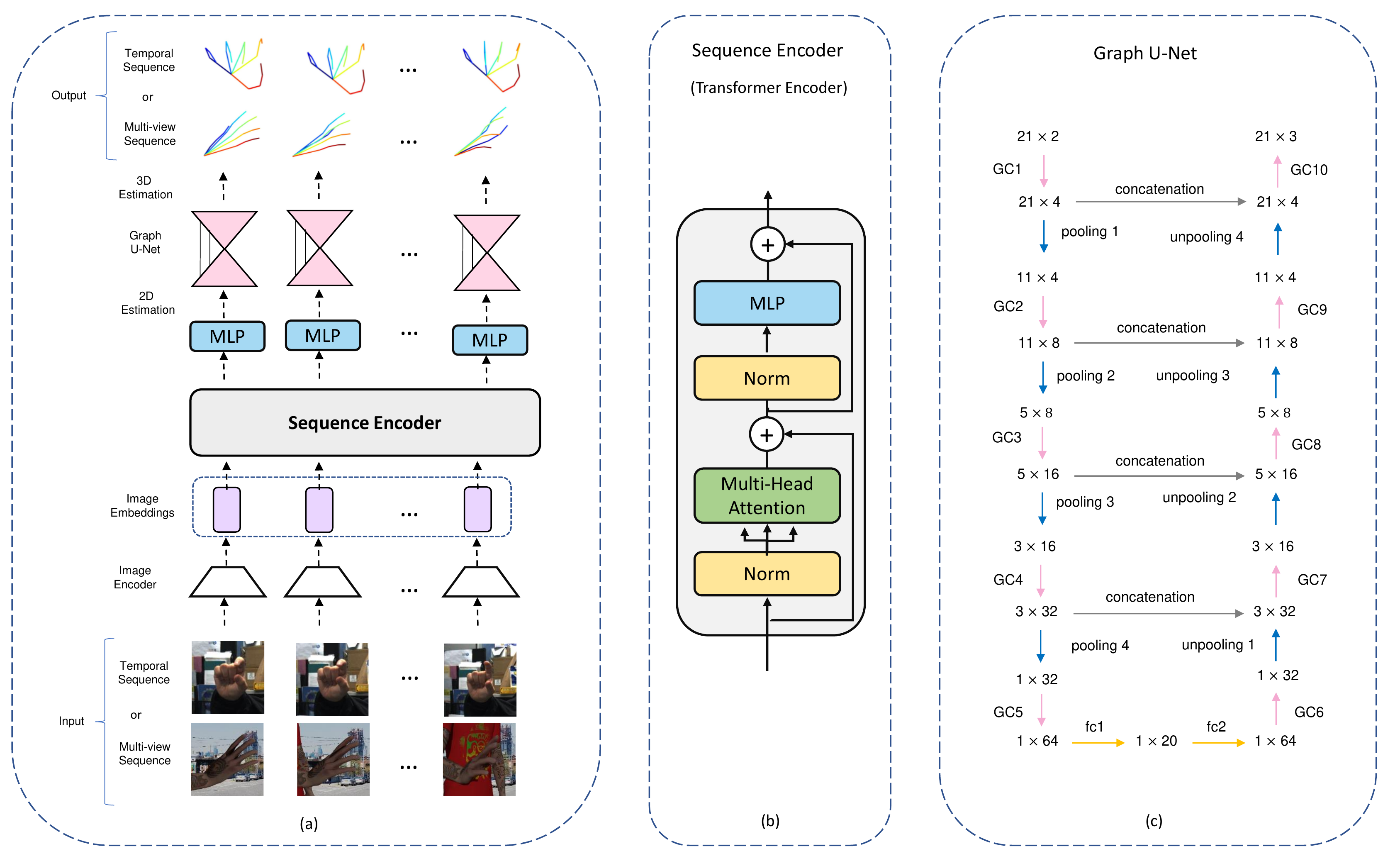}}
\caption {This diagram illustrates the pipeline for SeTHPose. An image encoder, a sequence encoder, and Graph U-Nets constitute the model. Transformer encoders take $N$ image embeddings in time or view from ResNet10. Afterward, an MLP that estimates 2D coordinates is applied to each encoder's output. In the next step, a Graph U-Net calculates the 3D hand pose from the corresponding 2D joint locations for each frame. }
\label{fig:model}
\end{figure*}

% \subsection{Model Overview}
Figure \ref{fig:model}(a) provides an overview of the proposed network. In our model, a ResNet is first used to encode the input images and produce individual hand frame embeddings. Then, inspired by recent works in other fields of computer vision that have used transformers to learn sequential information across time \cite{plizzari2021skeleton,zheng20213d} or space \cite{chen2021mvt,shuai2021adaptively}, we use a transformer encoder to learn the contextual sequential (temporal or angular) information. 
% We then use a transformer encoder which is powerful solutions for learning the Spatio-temporal \cite{plizzari2021skeleton,zheng20213d} and the Spatio-angular \cite{chen2021mvt,shuai2021adaptively} information which are practical for HPE \cite{khaleghi2021multi,simon2017hand, he2020epipolar,chen2021mvhm,yang2020seqhand, cai2019exploiting}. 
The output of the transformer is then used to produce 2D hand joint locations. SeTHPose then passes the estimated 2D joints  to a Graph U-Net to estimate the 3D hand poses. In the following sub-sections we discuss each of the main components of SeTHPose in detail.

% We follow the same idea of  2D to 3D lifting same as previous research \cite{iqbal2018hand,zimmermann2017learning,mueller2018ganerated,doosti2020hope} in HPE.

\subsection{Image Encoding}
In this research, we use a ResNet10 \cite{he2016deep} pre-trained on ImageNet \cite{russakovsky2015imagenet} to generate the image embeddings for each hand frame. Our encoder takes each hand frame $\phi_{i}$ and generates an image embedding \textcolor{black}{$x_{i}\in\mathbb{R}^{1 \times f}$}, where $i  = 1, 2, \cdots ,N$. By design, this encoder does not incorporate contextual sequential information because it only focuses on single frames. Thus, we need another component in our model to combine the individual embeddings in order to learn the surrounding context for each input frame. This leads to the use of a transformer encoder in SeTHPose, which we describe in the following sub-section.
% , and $N$ is the length of sequence,. 

\subsection{Sequence Encoding}
% Transformer Encoder developed for generating new word embedding based on comparing how much the word is relevant to its surrounding context. As a result, we consider hand sequence as a bag of words, where we could generate new feature embeddings which have been enriched by learning the relations between hand frames.
To exploit the sequence of embeddings obtained from individual frames and eventually generate contextually-informed poses, we use a transformer encoder.
% Given an input $X\in \mathbb{R}^{N \times f}$, for each image embedding $x_{i}\in\mathbb{R}^{1 \times f}$, linear transformations $w_q \in \mathbb{R}^{f \times d_q} $, $w_k\in \mathbb{R}^{f \times d_k} $, and $w_v \in \mathbb{R}^{f \times d_v} $, will map $x$ to queries, $q=xw_q$, keys $k=xw_k$ and, values $v=xw_v$. Here, $d_q$, $d_k$, and $d_v$, are the embedding dimensions.
Given an input $X\in \mathbb{R}^{N \times f}$, for each image embedding $x_{i}\in\mathbb{R}^{1 \times f}$, linear transformations $w_q \in \mathbb{R}^{f \times d_q} $, $w_k\in \mathbb{R}^{f \times d_k} $, and $w_v \in \mathbb{R}^{f \times d_v} $ are calculated. These values will map $x$ to queries $q=xw_q$, keys $k=xw_k$, and values $v=xw_v$. Here, $d_q$, $d_k$, and $d_v$, are the embedding dimensions.

To provide the relevancy of each image embedding $x_i$ based on other image embeddings available in the sequence $x_j$, $i,j= 1,2,\cdots, N$, we calculate relevancy score $s_{ij}= q_ik_i^T$. Next, by summing the multiplication of value vectors $v_j$ with $s_{ij}$, and normalizing the output with a softmax function, the new feature embedding for each image is calculated. This objective is named Scaled Dot-Product Attention \cite{vaswani2017attention} which is formulated as
\begin{equation}
\mathop{\rm Attention}(Q, K, V) = \mathop{\rm softmax}\left(\frac{QK^T}{\sqrt{d_k}}\right)V,
\end{equation}
% Here, $d_v$ acts as a scaling factor that helps during training by increasing the stability of the gradients through controlling small gradients while softmax is carrying large values. 
where, \textcolor{black}{$d_k$} acts as a scaling factor to improve gradient stability during training.

The use of multi-headed attention is a common strategy to improve the performance of the transformer where the attention mechanism is applied $H$ times with distinct learnable parameters according to
\begin{equation}
head_h = \mathop{\rm Attention}(Q W^Q_h, K W^K_h, V W^V_h),
\end{equation}% MultiHead(Q, K, V) = 
where $h = 1,2,\cdots,H$. The final result, $MH$, is calculated as
\begin{equation}
MH(Q, K, V) = \mathop{\rm concat}(head_1, ..., head_h)W_{out}.
\end{equation}

We utilize a multi-head self-attention mechanism to generate new hand image embeddings based on the sequential context (time or view) between the hand frames (see Figures \ref{fig:intro}(a) and \ref{fig:intro}(b) for examples). Ultimately, $N$ set of features $c_i \in \mathbb{R}^{{1} \times {L}}$, where \textcolor{black}{$L$} is the transformer's output feature size, is generated in this stage to be fed into an MLP to produce the corresponding 2D hand joint locations $z_i\in  \mathbb{R}^{21 \times 2}$, for $N$ hand frames.

% \subsection{Estimating 3D Poses from 2D Outputs}
\subsection{2D to 3D Pose Conversion}
As a next step, we convert the set generated 2D hand poses to 3D poses using a Graph U-Net, which consists of Graph Convolutional (GC) layers, as shown in Figure \ref{fig:model}(c). Since the hand skeleton consists of  a graph structure, the GC layers have been found to perform well for HPE \cite{doosti2020hope,Ge_2019_CVPR,cai2019exploiting,khaleghi2021multi,chen2021mvhm}.

For each GC layer, a graph $\textit{G} = ({K}, {A})$, with ${K}$ nodes and ${A} \in \mathbb{R}^{{K} \times {K}}$ as the adjacency matrix is defined. As per earlier research \cite{doosti2020hope,khaleghi2021multi}, we select a learnable adjacency matrix that has shown to achieve more effective results than predefined adjacency matrices. By feeding an input ${X} \in \mathbb{R}^{{K} \times {F}}$ to each GC layer, an output vector ${O} \in \mathbb{R}^{{K} \times {E}}$ is generated
\begin{equation}
\textit{O} = \sigma(\bar{{A}}{X}{W}),
\end{equation}
where $F$ and $E$ are the input and output feature sizes respectively, and ${W} \in \mathbb{F}^{{E} \times {L}}$ is the trainable weight matrix. $\bar{\textit{A}}$ is the normalized adjacency matrix \cite{kipf2016semi} of $G$, which is measured as
\begin{equation}
\bar{{A}} = {D}^{\frac{-1}{2}}\hat{{A}}{D}^{\frac{-1}{2}},
\end{equation}
where 
${D}$ is defined as the diagonal node degree matrix of $G$. Here $\hat{{A}}$ is calculated according to 
\begin{equation}
\hat{{A}} = {A} + {I},
\end{equation}
where ${I}$ is the identity matrix.

The Graph U-Net module is formed by multiple GC layers situated in a U-net encoder-decoder architecture with several skip connections between the corresponding layers. Finally, the Graph U-net converts each set of 2D hand joint locations $z_i\in  \mathbb{R}^{21 \times 2}$ to its corresponding 3D hand pose $p_i\in \mathbb{R}^{21 \times 3}$.

\subsection{Training and Loss Function}
We take two steps for training SeTHPose. First (Step 1), The pipeline is trained regardless of the sequence encoder. During this process, the transformer encoder is temporarily replaced by a fully connected layer. 
%and the $N$ dimension is moved inside the batch dimension. 
By doing so, we effectively implement HPE on every frame by optimizing the loss function
\begin{equation}
     L = \alpha L_{2D} + L_{3D},
\end{equation}
where $\alpha$ is set to 0.1, and $L_{2D}$ is defined as
\begin{equation}
    L_{2D} =||\hat{z} - z||_2,
\end{equation}
and $\hat{z}$ and $z$ are the predicted and ground truth 2D coordinates respectively. Similarly, the $L_{3D}$ is calculated according to 
\begin{equation}\label{eq:3d_loss}
    L_{3D} =||\hat{p} - p||_2,
\end{equation}
where $\hat{p}$ and $p$ are the predicted and ground truth 3D coordinates, respectively. 
% It should be noted that in this step, we use an image encoder that is already pre-trained on ImageNet. 
% as well as pre-training the U-Net graph structure. 
Following this, in Step 2, we freeze the image encoder and train the complete SeTHPose pipeline, effectively re-training the Graph U-Net and MLP, and training the sequence encoder (transformer encoder). Here, we use the loss function
\begin{equation}
    L = \frac{1}{N}\sum_{N}||\hat{p}_{i} - p_{i}||_2.
\end{equation}

\begin{table}[!t]
\caption{Performance comparison between SeTHPose and other methods on STB dataset.
} 
   \setlength
    \tabcolsep{3pt}
    \centering\begin{tabular}
    {| l | c |c |c |c|c|}
    \hline
    Method &Input Data  &$\downarrow$ Avg. EPE &$\uparrow$AUC (20-50)\\
    \hline
    \hline
    Zhang et al. \cite{zhang20163d} (PSO) &Image&-&0.709\\
    Zhang et al. \cite{zhang20163d} (ICPPSO) &Image&-&0.748\\
    Zhang et al. \cite{zhang20163d} (CHPR) &Image&-&0.839\\
    Panteleris et al. \cite{panteleris2018using} &Image&-&0.941\\
    Zimmerman \& Brox \cite{zimmermann2017learning} &Image&-&0.948\\
    Mueller et al. \cite{mueller2018ganerated} &Video&-&0.965\\
 
     Spurr et al. \cite{spurr2018cross} &Image&8.56&0.983\\
    Yang et al. \cite{yang2020seqhand} &Image&9.80&0.985\\ 
    Chen et al.\cite{chen2020dggan}&Image&9.11&0.990\\
     Yang et al. \cite{yang2019disentangling} &Image&8.66&0.991\\
    Li et al. \cite{li2020exploiting} &Image&-&0.996\\
   
      Iqbal et al. \cite{iqbal2018hand} &Image&-&0.994\\
    SeTHPose (Ours)  &Video&\textbf{7.841}&\textbf{0.998}\\
    \hline
    \end{tabular}
    \label{tab:STB}
\end{table}

\begin{figure}[!t]
\centerline{\includegraphics[width=0.45\textwidth]{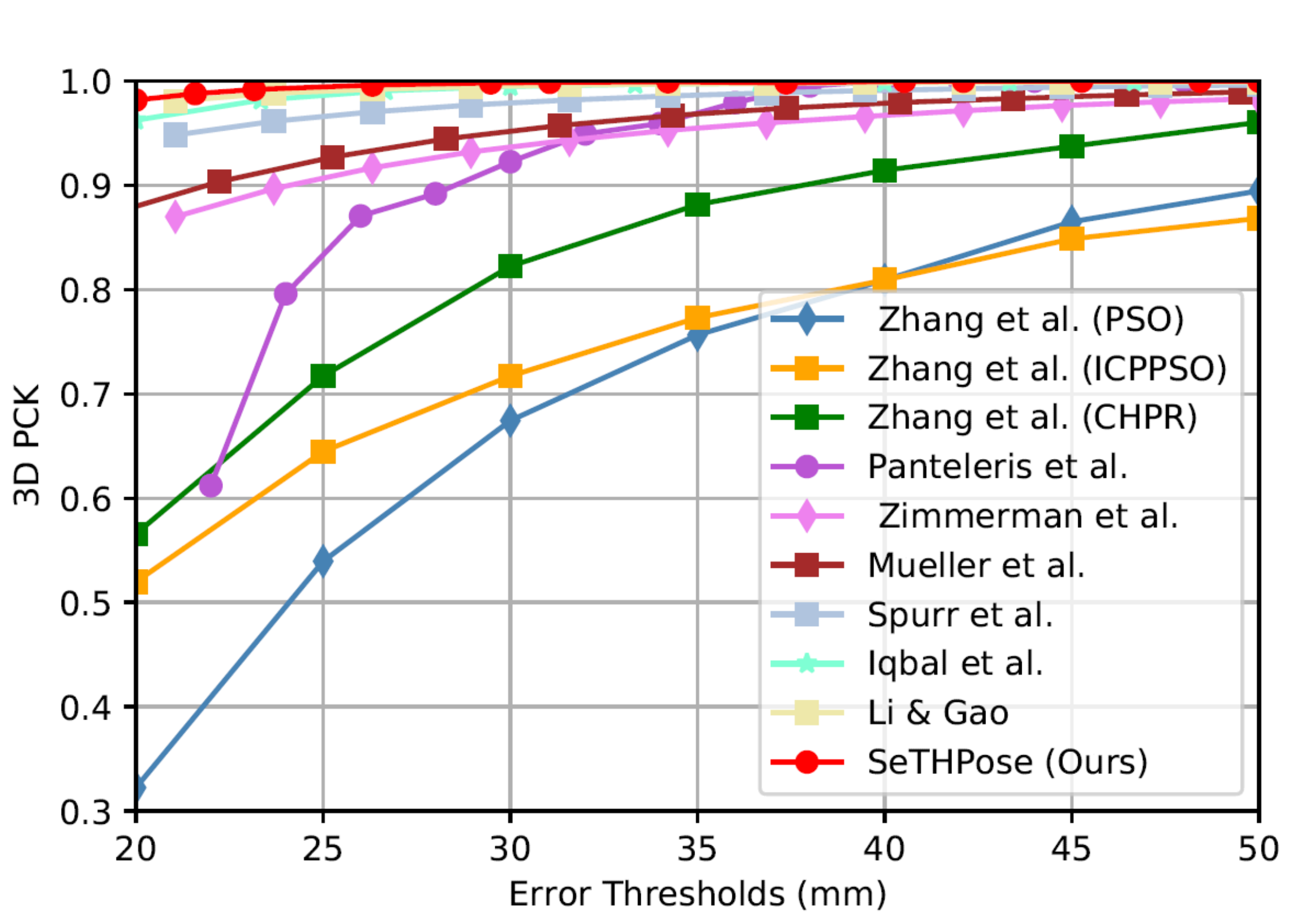}}
\caption {The comparison of 3D PCK curve of our proposed method against other solutions \cite{zhang20163d,panteleris2018using,zimmermann2017learning,mueller2018ganerated,spurr2018cross,iqbal2018hand,li2020exploiting} on STB dataset. }
\label{fig:STB_curve}
\end{figure}

\begin{figure}[!t]
\centerline{\includegraphics[width=0.85\columnwidth]{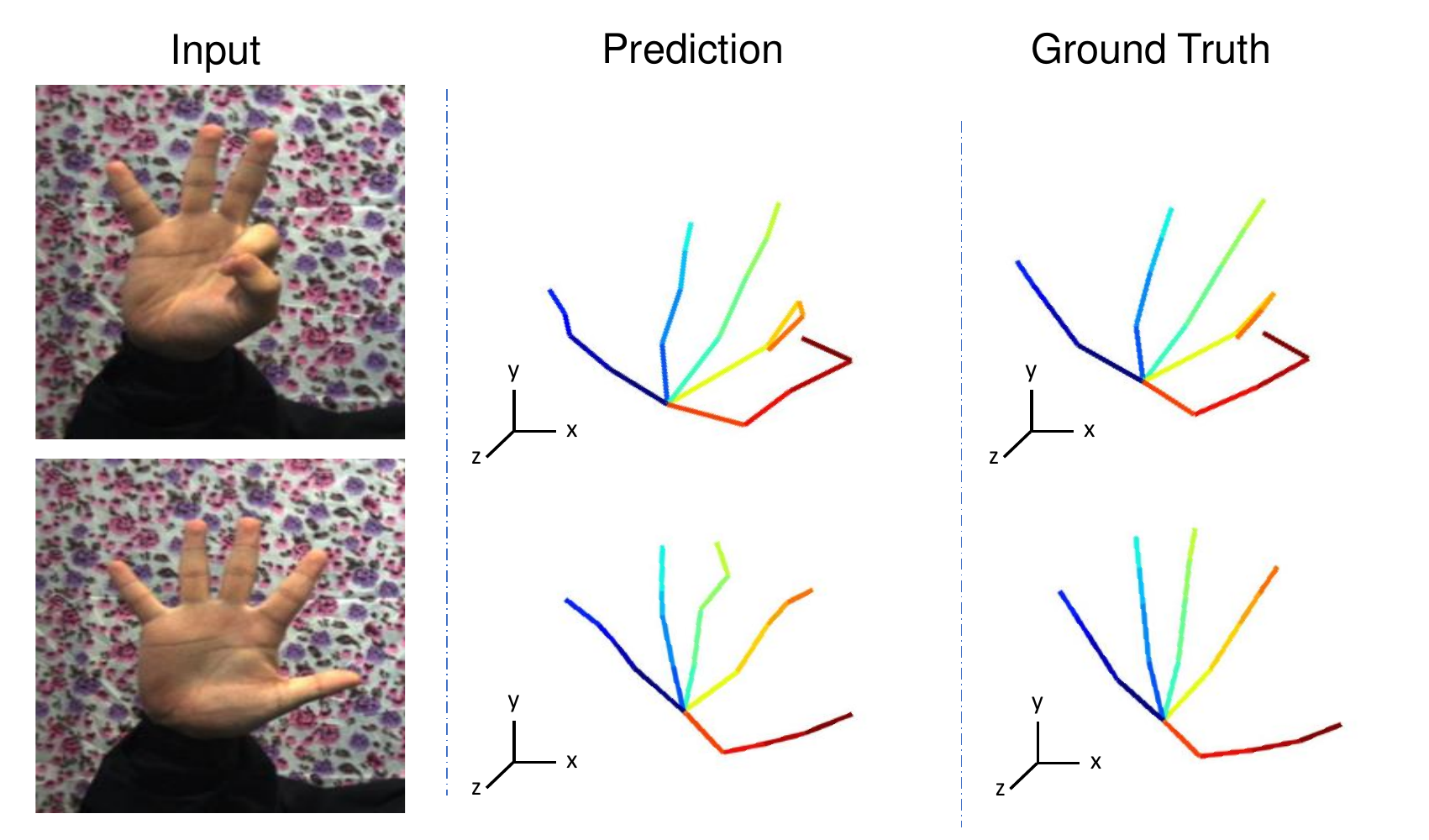}}
\caption {Sample visualization using SeTHPose for the STB dataset. }
\label{fig:stb_sample}
\end{figure}

\section{Experiments and Results}
% **
% In this section, we describe our experiments and report on the results. We also report on the outcome of ablation experiments and investigate the effects of various components of our network on the overall HPE performance.

\subsection{Datasets}
We use two publicly available 3D HPE datasets, STB \cite{zhang20163d} and MuViHand \cite{khaleghi2021multi}. We specifically select these datasets because they both contain hand poses in \textit{video} or \textit{multi-view} formats. Two other datasets, MVHM \cite{chen2021mvhm} and SeqHand \cite{yang2020seqhand}, which also contain multi-view and video-based data, could also have been used for this study, but the respective datasets were not made public at the time that we carried out our work. Next we describe the two datasets used in this study.

% is a new dataset that it has not been released yet, and SeqHand \cite{yang2020seqhand} which is a hand pose dataset in video format, does not released the data.}.

\begin{table}[!t]
\caption{Performance comparison between SeTHPose and other methods on MuViHand dataset with two different testing protocols.
}
  \setlength
    \tabcolsep{3pt}
    \centering\begin{tabular}
    {|c| l | c |c |c |c|}
    \hline
    Test & Method &Input Data &$\downarrow$ Avg. EPE  & $\uparrow$AUC \\
    \hline
    \hline
    {\multirow{8}{*}{\rotatebox[origin=c]{90}{cross-subject}}}
      &Boukhayma et al. \cite{boukhayma20193d}  &Image& 48.840   &0.280\\
     &Hasson et al. \cite{hasson2019learning} &Image&28.915  &0.574\\
     &Doosti et al. \cite{doosti2020hope} &Image& 18.895   & 0.634\\
     &Khaleghi et al. \cite{khaleghi2021multi} &Video &11.82 &0.766\\
      &Khaleghi et al.  \cite{khaleghi2021multi} &Mult-view Images &10.034  &0.808 \\
       &Khaleghi et al. \cite{khaleghi2021multi} &Mult-view Videos&8.881 &0.831\\
       &SeTHPose$_t$ (Ours)  &Video&7.697&0.846\\
      &SeTHPose$_v$ (Ours)  &Multi-view Images&\textbf{7.384}&\textbf{0.861}\\
    \hline
     {\multirow{8}{*}{\rotatebox[origin=c]{90}{cross-activity}}}
    &Boukhayma et al. \cite{boukhayma20193d}  &Image& 42.799   &0.287\\
     &Hasson et al. \cite{hasson2019learning} &Image&66.851  &0.152\\
     &Doosti et al. \cite{doosti2020hope} &Image& 46.745   & 0.217\\
      &Khaleghi et al. \cite{khaleghi2021multi} &Video&23.631 &0.557\\
       &Khaleghi et al. \cite{khaleghi2021multi} &Multi-view Images& 21.463&0.592\\
     &Khaleghi et al. \cite{khaleghi2021multi} &Mult-view Videos&20.375&0.608\\
       &SeTHPose$_t$ (Ours)  &Video&18.607& 0.638\\
      &SeTHPose$_v$ (Ours)  &Multi-view Images&\textbf{17.616} &\textbf{0.657}\\
    \hline
    \end{tabular}
    \label{tab:MuviHand}
\end{table}

\begin{figure}[!t]
\centerline{\includegraphics[width=0.92\columnwidth]{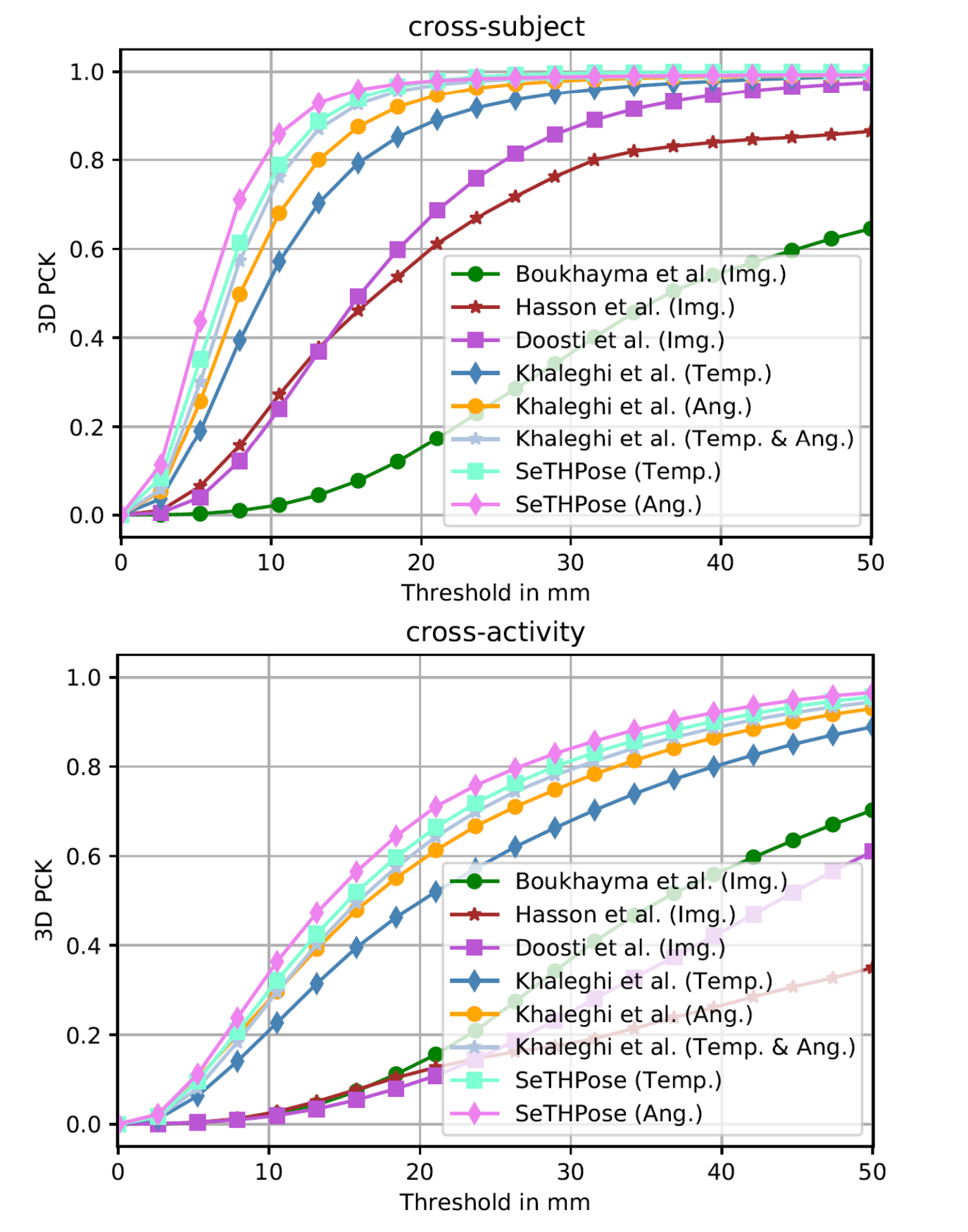}}
\caption {The comparison of 3D PCK curve of our proposed method against other solutions \cite{boukhayma20193d,hasson2019learning,doosti2020hope,khaleghi2021multi} on two testing protocols of MuViHand dataset; top: cross-subject and bottom: cross-activity.}
\label{fig:MuViHand_curve}
\end{figure}

\textbf{Stereo Hand Pose Tracking Benchmark (STB)} \cite{zhang20163d} is a single-view real-world dataset that captures 12 hand \textit{\textbf{videos}}. 
% as only one subject performs either counting or random hand poses.
We use the same training/test splits as previous studies \cite{boukhayma20193d,spurr2018cross,chen2020dggan,yang2020seqhand}, training the model on ten videos and testing it on the remaining videos.

% Similar to previous studies \cite{boukhayma20193d,spurr2018cross,chen2020dggan,yang2020seqhand} 
% we use the same training/test splits, training the model on ten videos and testing it on the remaining videos.

\textbf{Multi-view video based Hand (MuViHand)} \cite{khaleghi2021multi} is a synthetic dataset containing video footage of 10 subjects performing 17 activities from multiple perspectives. In this study, we use the same train/test protocol as in the original paper. Specifically, in the cross-subject protocol, we train on subjects $3,4, \cdots, 9$ and test on subjects $ 1, 2, 10$, while in the cross-activity protocol, we train on activities $1,2, \cdots, 7, 9,10, \cdots, 18$, and test on activities $8, 19$.

\subsection{Metrics}
Following the standard protocols used for HPE\cite{zimmermann2017learning}, we calculate three metrics for evaluating our method. These metrics are (\textit{i}) mean endpoint error (EPE), (\textit{ii}) the proportion of correct keypoints whose Euclidean distance error is less than a threshold (3D PCK), and (\textit{iii}) the area under the curve of the 3D PCK (AUC). It should be noted that the threshold for 3D PCK is selected similar to previous works in the area \cite{zimmermann2017learning,iqbal2018hand,spurr2018cross,mueller2018ganerated,panteleris2018using,li2020exploiting,zhang20163d,khaleghi2021multi}.

% which has been calculated with threshold between 0-50 mm for 
% with threshold between 0-50 mm, 

% aslo for a fair comparison with other methods on STB dataset the PCK threshold between 20-50 mm have been calculated, 

\subsection{Implementation Details} \label{sec:Implementation}

The two datasets used in this study contain different types of data. STB is a real dataset containing temporal sequences, while MuViHand is a synthetic dataset containing both temporal and angular sequences. Thus, the training parameters used for the two datasets are different to accommodate for the varying levels of difficulty and synthetic versus real nature of the data. We implement our model using PyTorch and perform the training on an Nvidia GeForce GTX 2070 Ti GPU.

% As mentioned earlier, the \textit{STB} dataset only contains temporal contextual data (videos).
For training with the STB dataset, in Step 1, we use an initial learning rate of 0.005, and decrease it by a factor of 0.09 every 50 epochs for 500 epochs. Following this, in Step 2, we train SeTHPose with $N =$~5 for 5000 iterations at an initial learning rate of 0.001, multiplied by 0.95 every 50 iterations.

\begin{figure}[!t]
\centerline{\includegraphics[width=0.75\columnwidth]{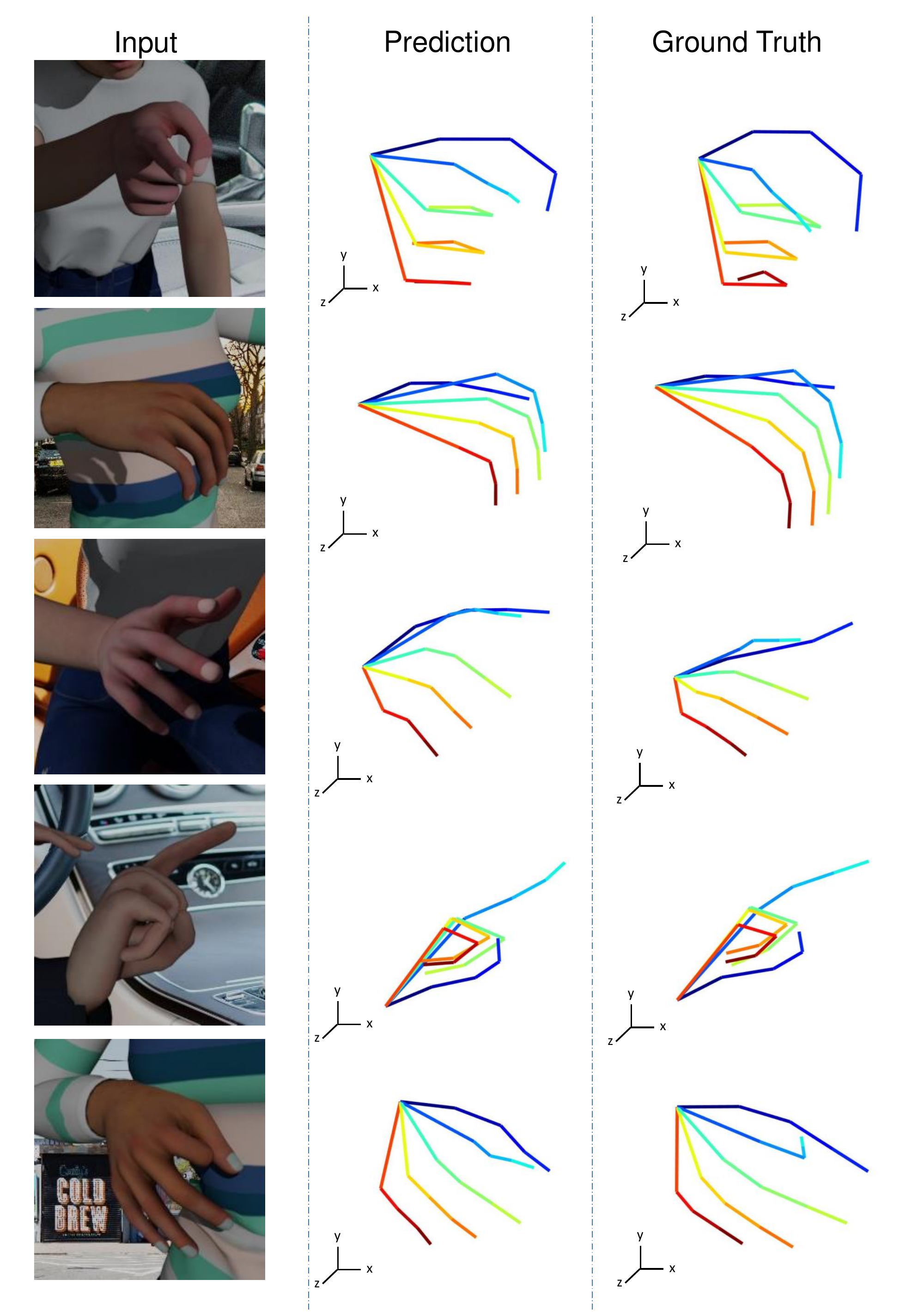}}
\caption {Sample visualization using SeTHPose for the MuViHand dataset. }
\label{fig:MuViSample}
\end{figure}

Given that MuViHand dataset contains both temporal (video) data and multi-view images, we explore both temporal and angular relationships using our model. Accordingly we have SeTHPose$_t$ and SeTHPose$_v$, which refers to two separate experiments, one for learning temporal relationships and the other for learning angular relationships. In Step 1 of training, we train both variations for 500 epochs at an initial learning rate of 0.001, multiplied by 0.1 after every 100 epochs. Afterwards, SeTHPose$_t$ is trained with $N=$~5 while SeTHPose$_v$ is trained with $N=$~3 with an initial learning rate of 0.001, is multiplied by 0.9 every 100 epochs for 350 epochs.

% As a part of our implementation, we have centralized the hands for all the images and resized them to $224 \times 224$ aligned with ResNet-10 input size. 

\subsection{Performance and Discussion}
We evaluate and compare the performance of SeTHPose on STB dataset against other state-of-the-art single-image HPE methods \cite{zimmermann2017learning,iqbal2018hand,spurr2018cross,mueller2018ganerated,panteleris2018using,chen2020dggan,yang2019disentangling,li2020exploiting,zhang20163d} and the only video-based method available in the literature on this particular dataset \cite{yang2020seqhand}. We present this comparison in Table \ref{tab:STB}. The performance values have been obtained from the original papers (in some cases the EPE metric has not been reported). SeTHPose shows excellent performance on STB with no additional training data in contrast to \cite{chen2020dggan,yang2019disentangling,yang2020seqhand}, which rely on training on several datasets for better generalization. Furthermore, in Figure \ref{fig:STB_curve}, we present the 3D PCK curves of our method in comparison to \cite{zimmermann2017learning,iqbal2018hand,spurr2018cross,mueller2018ganerated,panteleris2018using,li2020exploiting,zhang20163d} for different thresholds (20-50). Also, in Figure \ref{fig:stb_sample}, we illustrate several samples of input images and output poses estimated by SeTHPose.

We present the results of SeTHPose (both SeTHPose$_t$ and SeTHPose$_v$) on MuViHand dataset and compare the performances against \cite{doosti2020hope,hasson2019learning,boukhayma20193d} as well as two additional variations of \cite{khaleghi2021multi}. 
% In the table, SeTHPose$_t$ and SeTHPose$_v$ denote have shown robust performance.
Interestingly, both variations of our method outperform all other existing methods on this dataset, including both multi-view and video-based methods.
% the variant \cite{khaleghi2021multi}, which utilizes both temporal and angular input data simultaneously. With this example, you can clearly see how useful the transformer is for HPE tasks compared to other recurrent neural layers, which result in comparable results with fewer data. 
A graph showing the PCK curves for thresholds 0 to 50 for our method and other methods is presented in Figure \ref{fig:MuViHand_curve} for both test protocols. Also, in Figure \ref{fig:MuViSample}, we illustrate multiple examples of our network's performance on the MuViHand dataset. 

To evaluate the sensitivity of our method against the number of attention heads $H$, we repeat the experiments with different variations of the transformer-based sequence encoder by setting $H$ to 1, 2, 4, 8, and 16.
% A study that changed the number of head attention (1, 2, 4, 8, 16) in the SeTHPose in both datasets is carried out to evaluate the impact of this parameter. 
We present the results in Table\ref{tab:head}, where we observe that 4, 8, and 16 attention heads achieve relatively close results. Yet, $H=$~8 provides slightly better overall performance, which was selected for our model.
% different reul lend themselves to different types of data. 

\begin{table}[!t]
\caption{Impact of $H$ which is used in the transformer encoder in the SeTHPose on the EPE. 
}
  \setlength
    \tabcolsep{3.5pt}
    \centering\begin{tabular}
    {| c |c|c| c |c |c |c|c|}
    \hline
     \multicolumn{2}{|c|}{Dataset}&\backslashbox{Method}{$H$}&1 &2 &4 &8&16 \\
    \hline
    \hline
    \multicolumn{2}{|c|}{STB}
     &SeTHPose&10.034 &9.406& 9.069&7.841&8.787 \\\cline{1-8}
     {\multirow{4}{*}{\rotatebox[origin=c]{90}{MuVi.}}}
     &{\multirow{2}{*}{\rotatebox[origin=c]{90}{sub.}}}
    &SeTHPose$_t$ &8.878 &8.328 &8.159 &7.697 & 7.825\\
    &&SeTHPose$_v$ & 8.504 &7.842 &7.264&7.384& 7.805 \\\cline{2-8}
  
     &{\multirow{2}{*}{\rotatebox[origin=c]{90}{act.}}}
      
      &SeTHPose$_t$ &19.384 &18.648&19.424 &18.607&19.171 \\
    &&SeTHPose$_v$ & 20.043  &17.287 & 17.837&17.616&16.708 \\\cline{2-8}
    \hline
    \end{tabular}
    \label{tab:head}
\end{table}

\begin{table}[!t]
\caption{Ablation results on SeTHPose.
}
  \setlength
    \tabcolsep{3pt}
    \centering\begin{tabular}
    {| c |c| c |c |c |c|c|}
    \hline
     \multicolumn{2}{|c|}{Dataset}&Method &Temporal &Angular &$\downarrow$ Avg. EPE  & $\uparrow$AUC \\
    \hline
    \hline
    \multicolumn{2}{|c|}{\multirow{2}{*}{\rotatebox[origin=c]{0}{STB}}}
    &\textcolor{black}{Ablation}&\xmark & \xmark  &32.450  &0.600\\
    \multicolumn{2}{|c|}{}
    &SeTHPose&\cmark & \xmark&7.841&0.998\\\cline{1-7}
     {\multirow{6}{*}{\rotatebox[origin=c]{90}{MuViHand}}}
     &{\multirow{3}{*}{\rotatebox[origin=c]{90}{cr.-sub.}}}
    &\textcolor{black}{Ablation} &\xmark & \xmark  &14.529  &0.723\\
    &&SeTHPose$_t$  &\cmark & \xmark&7.697&0.846\\
    &&SeTHPose$_v$   &\xmark & \cmark &7.384&0.861\\\cline{2-7}
  
     &{\multirow{3}{*}{\rotatebox[origin=c]{90}{cr.-act.}}}
      &Ablation &\xmark & \xmark &44.021 &0.246 \\
      &&SeTHPose$_t$  &\cmark & \xmark&18.607& 0.638\\
       &&SeTHPose$_v$  &\xmark & \cmark&16.708 &0.657\\
    \hline
    \end{tabular}
    \label{tab:abliation}
\end{table}

\subsection{Ablation Study}
To explore the impact of the transformer encoders on learning the sequential contexts in our method, we perform ablation studies on SeTHPose. To this end we remove the transformer structure from the pipeline, thus estimating the final pose from a single image. Table \ref{tab:abliation} presents the results of this experiment for both datasets, STB and MuViHand, with both testing protocols (cross-subject and cross-activity). It can be observed that the performance of our method drops in the absence of learning contextual information, indicating that temporal and multi-view sequences provide additional valuable data points which can be effectively and successfully exploited by the transformer in our model. In particular, the EPE increases by around 25 \textit{mm} for the STB dataset, and by an average of 7 \textit{mm} and 27 \textit{mm} for MuViHand dataset in the cross-subject and cross-activity schemes, respectively.

\section{Conclusion}
We present SeTHPose, a model capable of exploiting sequential hand data (video or multi-view images) using transformers to perform accurate 3D HPE. Our model includes multiple components to encode individual hand frames, learn the sequential contexts, and estimate 3D hand poses. Our model is evaluated by a number of experiments, which include comparisons with state-ofthe-art methods using publicly available datasets and ablated baselines. These experiments demonstrate the effectiveness of our method by outperforming other solutions in the area.%, while also proving its 

\bibliographystyle{unsrt}  
\bibliography{IEEE}

\ifCLASSOPTIONcaptionsoff
  \newpage
\fi

\end{document}